# Impact of an Autonomous Shuttle Service on Urban Road Capacity: Experiments by Microscopic Traffic Simulation

**Sudipta Roy**
Department of Civil, Environmental, and Construction Engineering
University of Central Florida,
Orlando, Florida, United States, 32816
Email: sudipta.roy@ucf.edu
ORCID: 0000-0002-9588-6013

**Bat-hen Nahmias-Biran**
Department of Civil Engineering
Ariel University, Ariel 40700, Israel
and
Department of Civil and Environmental Engineering
Massachusetts Institute of Technology,
Cambridge, Massachusetts, United States, 02139
Email: bathennb@ariel.ac.il
ORCID: 0000-0002-3223-4894

**Samiul Hasan**
Department of Civil, Environmental, and Construction Engineering
University of Central Florida,
Orlando, Florida, United States, 32816
Email: samiul.hasan@ucf.edu
ORCID: 0000-0002-5828-3352

## ABSTRACT

Autonomous vehicles are expected to transform transportation systems with rapid technological advancement. Human mobility would become more accessible and safer with the emergence of driverless vehicles. To this end, autonomous shuttle services are currently introduced in different urban conditions throughout the world. As a result, studies are needed to assess the safety and mobility performance of such autonomous shuttle services. However, calibrating the movement of autonomous shuttles in a simulation environment has been a difficult task due to the absence of any real-world data. This study aims to calibrate autonomous shuttles in a microscopic traffic simulation model and consequently assess the impact of the shuttle service on urban road capacity through simulation experiments. For this analysis, a prototype of an operational shuttle system at Lake Nona, Orlando, Florida is emulated in a microscopic traffic simulator during different times of the day. The movements of autonomous vehicles are calibrated using real-world trajectory data which help replicate the driving behavior of the shuttle in the simulation. The analysis reveals that with increasing frequency of the shuttle service the delay time percentage of the shared road sections increases and traveling speed decreases. It is also found that increasing the speed of shuttles up to 5 mph during off-peak hours and 10 mph during peak hours will improve traffic conditions. The findings from this study will assist policymakers and transportation agencies to revise policies for deploying autonomous shuttles and for planning road infrastructures for shared road-use of autonomous shuttles and human driven vehicles.

## INTRODUCTION

Vehicular automation has created opportunities to reshape urban mobility landscape and transportation systems. Autonomous vehicles (AV) or connected autonomous vehicles (CAV) are expected to serve a significant portion of travel demand in future. Increasing penetration of autonomous vehicles is expected to significantly contribute to—crash reduction, travel time reduction, fuel savings, and equity improvement—which are attracting policymakers to introduce autonomous vehicles in different cities (*1*–*6*).

The potential impact of vehicle automation is not limited to personal transportation only; public transit is also expecting automation to improve safety and efficiency. To this end, autonomous shuttles have emerged as a highly promising transportation mode, with several pilots being introduced in different places in the last few years (*7*). Some of the busiest cities of the world including Singapore, Paris, Tokyo, and Perth and several states including Florida, California, and Nevada in the USA are implementing autonomous shuttles in different urban contexts (*8*–*12*). They represent a prospective solution for both on-demand and scheduled mass transit, offering significant potential for future of public transportation (*13*–*15*).

One of the most prominent features of autonomous vehicles is that these vehicles can be operated with a smaller headway than usual—increasing roadway capacity and causing less congestion and reduced travel time (*16*). But some studies also suggest that in several cases automated vehicles are deteriorating road capacity as an effect of low-level automation or shorter stop-to-stop gaps (*17*). Cao and Ceder (*18*) found that shuttle systems offering demand-based skip-stop services provide a better service to passengers due to reduced travel time, with improved network performance by reducing congestion. For autonomous shuttles, network performance is affected by operating conditions like use of a dedicated lane and increasing penetration of autonomous vehicles (*19*).

As currently autonomous shuttles are being piloted in many communities considering the opportunities it provides for mobility without vehicle ownership, there are growing interests among policymakers and traffic management agencies to know how these shuttles would impact roadway capacity. Although several studies have assessed the impact of autonomous shuttle movement, those studies have some limitations. Mixed traffic, comprising autonomous shuttles and the other types of vehicles, creates unique traffic flow characteristics which are comparatively new to researchers. This requires a proper calibration of vehicle movements in simulation platforms.

In this study, we aim to answer the research question: what are the impacts of autonomous shuttle movement on road capacity in an urban traffic network? The objective of this study is to evaluate the impact of autonomous shuttles on urban road environment by adopting a microscopic traffic simulation approach. To accomplish this objective, we have created an autonomous shuttle movement network in a microscopic traffic simulation platform following a real-world deployment of autonomous shuttles in Lake Nona, Orlando, Florida. The simulation model has been calibrated for traffic demand and vehicle movement parameters for both autonomous shuttles and human-driven vehicles (HDVs) using detector data and real-world vehicle trajectory data obtained from field measurements. Later, we have performed impact analysis of autonomous shuttle movement under different scenarios including travel demand projections and possible headways of the operated autonomous shuttles.

The calibration of simulation parameters for autonomous vehicles and/or shuttles was not possible in previous research due to the lack of real-world data. However, automated shuttles are being piloted in different urban environments. These operations are providing opportunities to researchers to calibrate simulation parameters of autonomous shuttles incorporating real-world travel demand and vehicle movement data to produce more accurate results of potential impact of autonomous shuttles. The novelty of this study is that we calibrated the vehicle movement of autonomous shuttles using vehicle trajectory data collected through field measurements from a real-world deployment of autonomous shuttles.

There are significant implications of this study. This work will provide an opportunity for traffic management agencies to have a better understanding of the impact of operating autonomous shuttles on urban roads in mixed traffic conditions. The calibrated car-following and lane-changing behaviors of autonomous shuttles and HDVs will help to correctly emulate traffic movement in a real-world environment. This study can demonstrate the real-world impact of autonomous shuttles on road capacity and suggest policies for shuttle operations to optimize travel time and traffic flows for autonomous shuttles and other road users.

## STATE-OF-THE-ART REVIEW

The review on the state-of-the-art focuses on previous studies on the impacts of autonomous vehicles/shuttles on roadway capacity, the modeling procedures adopted by existing studies, typical scenarios modeled for impact evaluation, and the calibration procedure of microscopic traffic simulation models.

Several studies have investigated the impact of autonomous vehicles (AVs) or connected autonomous vehicles (CAVs) carrying passengers on transportation network. Most of the impact assessments used microscopic traffic simulation models using open-source (e.g., SUMO) or commercial (e.g., VISSIM and AIMSUN) simulation platforms. These studies revealed that the penetration rate of AVs is an important parameter on how AVs can influence road capacity. Using VISSIM, Park et al. (*20*) observed an improvement in traffic flow capacity with increasing AV penetration rate in an urban road network. This study suggests that for saturated penetration level of AVs, road network capacity can increase up to 40%. Using microscopic simulation on freeway, Talebpour and Mahmassani (*21*) showed that AVs can significantly increase highway capacity using a reserved lane when the penetration rate is more than 50% for two-lane freeways and more than 30% for four-lane freeways. Lu et al. (*16*) used microscopic simulation on SUMO to assess how different penetration levels of AVs impact urban road network and found similar correlations between AV penetration rates and roadway capacity.

Some studies also focused on the impact of autonomous shuttles depending on their operating routes. Ziakopoulos et al. (*19*) studied the impact of on-demand autonomous shuttle on urban road network via microscopic simulation considering different scenarios such as mixed movement of autonomous shuttles and human driven vehicles (HDVs) and a dedicated lane for shuttle movement. This study also considered different penetration rate of autonomous shuttles and found that a higher penetration rate of autonomous shuttles reduces travel time. Similarly, Oikonomou et al. (*22*) studied the impact of autonomous shuttles on road capacity, traffic safety, and environment through microscopic simulation where they showed that operations of a shuttle service increased delay, specifically during off-peak periods. Mourtakos et al. (*23*) studied the impact of on-demand mobility services through simulation where they found that implementation of the services would lead to decreased delay at a network level.

The reviewed literature revealed some gaps in the existing works on the impact analysis of autonomous shuttle on urban road capacity. All works on impact analysis of autonomous shuttles are based on simulation. However, car-following models simulating the behavior of autonomous shuttles and human driven vehicles (HDVs) were not calibrated in earlier works due to the lack of real-world mixed road use data observing the interactions between automated shuttles and HDVs. Since autonomous shuttle operations are comparatively new, no vehicle trajectory data is currently publicly available to the best of our knowledge.

The purpose of calibrating a simulation model is to emulate the characteristics of real-world traffic by tuning different simulation parameters and reducing errors between field observations and simulated data. The calibration of traffic simulation models is conducted in two different levels. The first level is to calibrate the volume of traffic or the origin-destination demand matrix and the route choice behavior model and the other level is calibrating the car-following and lane-changing behaviors used in simulation (*24*, *25*). Both levels are important for a simulation-based approach to replicate a real-world scenario. Different sources of real-world data are used for the calibration process. For instance, data from vehicle detectors, GPS trajectories from vehicles, and video-based trajectories are used for simulation calibration. Detector data reveal macroscopic properties and vehicle trajectory data reveal microscopic properties (*26*, *27*). Several goodness-of-fit metrics such as Geoffrey E. Havers (GEH) statistics, root mean square error (RMSE), and mean absolute percentage error (MAPE) are typically used in the calibration process to quantify calibration errors (*24*–*28*).

Different car-following models are used for modeling AVs and HDVs in various circumstances. The car-following models use different parameters such as maximum and minimum desired acceleration, deceleration values, desired speed, reaction time, sensitivity etc. to replicate the real-world road traffic environment in simulation (*29*, *30*). The car-following models used in different studies include General Motors (GM) model, collision avoidance models (e.g., Gipps model), psychophysical models (e.g., Wiedemann model), the optimal velocity model, the intelligent driver model (IDM), the adaptive cruise control (ACC), and cooperative adaptive cruise control (CACC) models (*29*, *31*). The calibration of these car-following models and consequent lane-changing behaviors are done by comparing actual and simulated characteristics like distance, speed, headway, and lane-changing tendencies using aggregated count data or vehicle trajectory data (*24*, *25*, *27*, *29*, *30*, *32–35*). Due to the lack of real-world data on operations of AVs, previous studies have calibrated their simulation using aggregated volume data (*19*, *22*, *23*).

## METHODOLOGY

In this section, we present the study area, models for autonomous shuttle movement, the calibration framework, and the scenarios developed to compare the impact of shuttle movement.

### Study Area with a Real-world Autonomous Shuttle System

Autonomous mobility service provider Beep is operating shuttle services in Lake Nona region of Orlando, Florida (see **Figure 1**). Lake Nona is a mixed urban area close to Orlando International Airport (*8*). Beep is operating daily its current shuttle services on six routes in limited time periods. The current Beep movement is on-demand, and the shuttle is autonomous in only two routes; but the organization is planning to expand its autonomous shuttle service to other routes. All six operating routes with route length, number of stops, and current conditions are provided in **Table 1**.

**TABLE 1 Beep autonomous shuttle movement routes**

| No. | Routes | Length (miles) | Number of stops | Currently Autonomous (Yes/No) |
|---|---|---|---|---|
| 1 | Wave - Canvas | 2.19 | 2 | Yes |
| 2 | Wave - LNPC | 0.56 | 2 | Yes |
| 3 | Heroes Park | 2.29 | 2 | No |
| 4 | VA - Disco Dog | 1.16 | 3 | No |
| 5 | Medical City | 1.77 | 2 | No |
| 6 | Springhill - Boxi Park | 4.96 | 2 | No |

The simulation model is developed in Aimsun Next (version 8.3.1) platform, a commercial software package for transportation modeling and planning through simulation (*36*). As microscopic behaviors such as car-following and lane-changing features are included in our research scope, a microscopic simulation platform is selected for this study. We have included the main road network of the study area in our simulation. The road network geometry is first imported from OpenStreetMap platform (*37*) and then the geometric and functional characteristics of the road sections such as the number of lanes, directions, free flow speeds, and capacities are manually corrected to replicate the actual road network in Lake Nona. The intersection characteristics such as the number of allowed movements, stop, yield directions and signal plans are provided in the model. All six operating or probable routes of the shuttle movement are also incorporated in the simulation (see **Figure 1**).

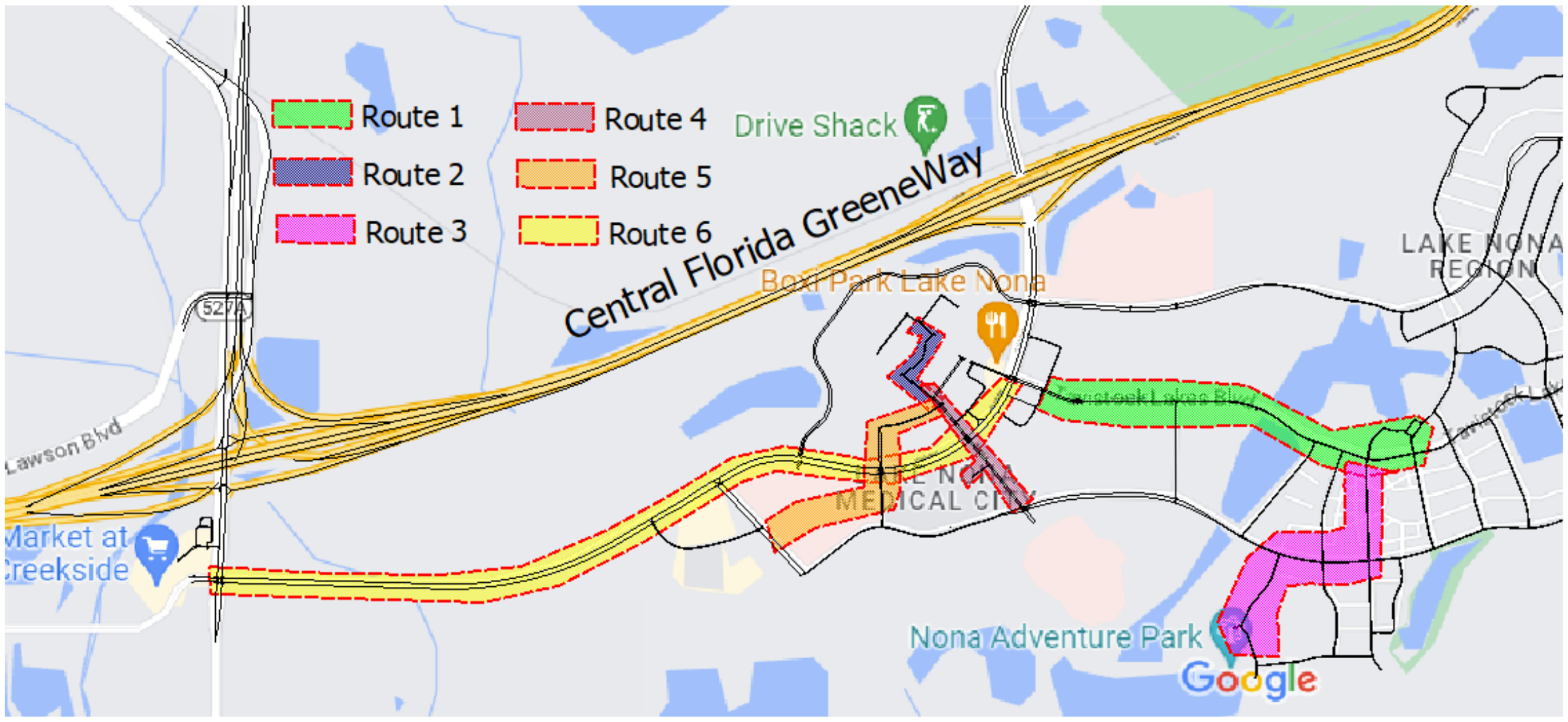

**Figure 1 Shuttle Routes in Lake Nona Region simulation (using Google map background)**

Two types of vehicles are simulated including passenger cars as human driven vehicles (HDVs) and Beep autonomous shuttles. For passenger cars, the vehicle dimension is provided as the standard dimension of a sedan; and for modeling Beep shuttles, the dimensions of the shuttles are obtained from their manufacturer company Navya (*38*). The shuttle stops are provided as they are currently located. Pedestrian movement is not included in the simulation as the number of pedestrians is quite negligible.

**Vehicle Movement Calibration**

The vehicle movements in simulations are mainly controlled by their speed and their dynamic and microscopic models which are used to accurately represent and imitate driving behaviors. **Figure 2** presents the vehicle parameters included in the simulation platform for autonomous shuttle and HDV movements through the combination of different models.

Gipps car-following and lane-changing models are applied in previous studies for both HDV and AV/CAV modeling and it is chosen as the car-following and lane-changing model in this study (*19*, *22*, *23*, *29*). The reasons for choosing these models are because these can emulate the behavior of traffic flow. This model predicts the behavior of the following vehicle by assuming that drivers have preferred limit for the vehicle acceleration and deceleration for collision avoidance and safe driving which is applicable for both autonomous shuttles and HDVs (*29*, *39*). The vehicle movement models are calibrated using the vehicle trajectories obtained by field data collection efforts.

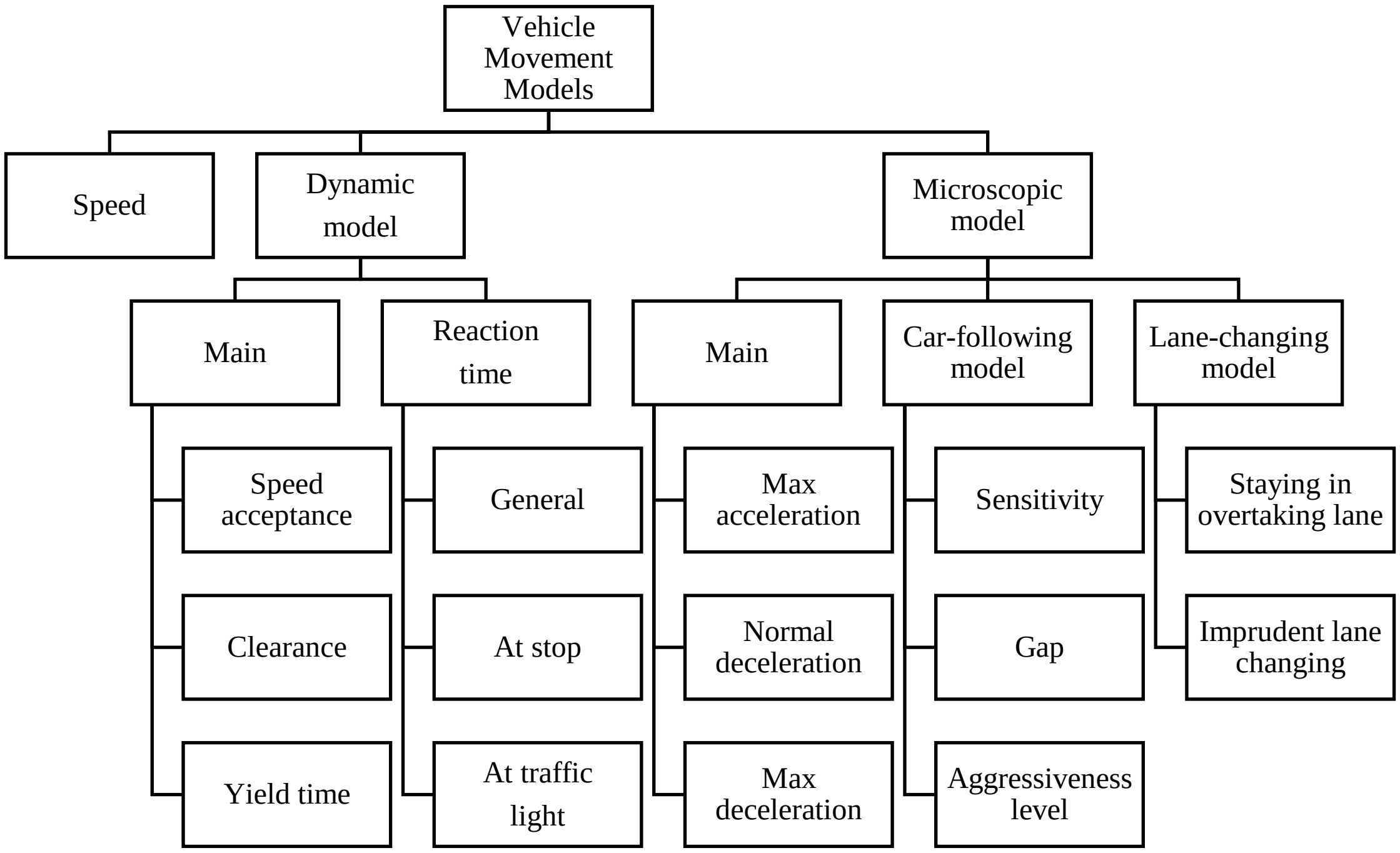

**Figure 2 Vehicle parameters included in the microscopic simulation**

The field data collection efforts and calibration procedures are described as follows:

1. We have collected vehicle trajectory data by keeping a GPS device in a sedan car (to reflect an HDV) and in the autonomous shuttles (**Figure 3**). The devices can collect data on minimum 1 second interval, with 10 Hz frequency (*40*). Thus, the trajectory data are obtained in 1-second intervals for both vehicle types. The numbers of trips taken for trajectory data collection are given in **Table 2**.
2. We collected data from different segments of three routes (route nos. 1, 3, and 6) which are comparatively longer with higher traffic volumes. A total of 22 road segments (6 segments for route no. 1, 6 segments for route no. 3, and 10 segments for route no. 6) are considered.
3. Trajectory data from the autonomous shuttle is collected on route 1, where we ride the bus with the GPS device and completed one full trip. As the shuttle is operated visibly at a very low speed, effectively it does not follow any HDV in its route. But sometimes it takes unscheduled stops to release the queue of HDVs behind it. This stopping time is ignored in our calculation. So, the free-flow speed of the shuttle is traced in our trajectory data.
4. The travel times for each road segment are calculated later from the obtained trajectory data and then averaged to get the mean travel time (in case of multiple trips) for all of the three conditions (HDV following the shuttle, HDV without following the shuttle, and shuttle movement) mentioned in **Table 2**.
5. Similar scenarios are developed in the simulation to observe the travel time of vehicles. Different values for vehicle modeling parameters (shown in **Figure 2)** are used to generate different combinations of vehicle model parameters for use in simulation. The initial values for the trial-and-error process are obtained from relevant literature (*19*, *22*, *31*). The conditions in **Table 2** are observed in the simulation and the mean travel time of the simulated vehicles are calculated for these conditions. The observed free movement travel time of HDVs and shuttles from trajectory data in different road segments and the simulated trajectory data are compared to get the accuracy of the vehicle models. The mean absolute percentage error (MAPE) value of the travel time is used to evaluate the simulation parameter combinations. The parameter combinations with the least MAPE value are finally selected for modeling HDVs and autonomous shuttles separately (see **Table 3)**.

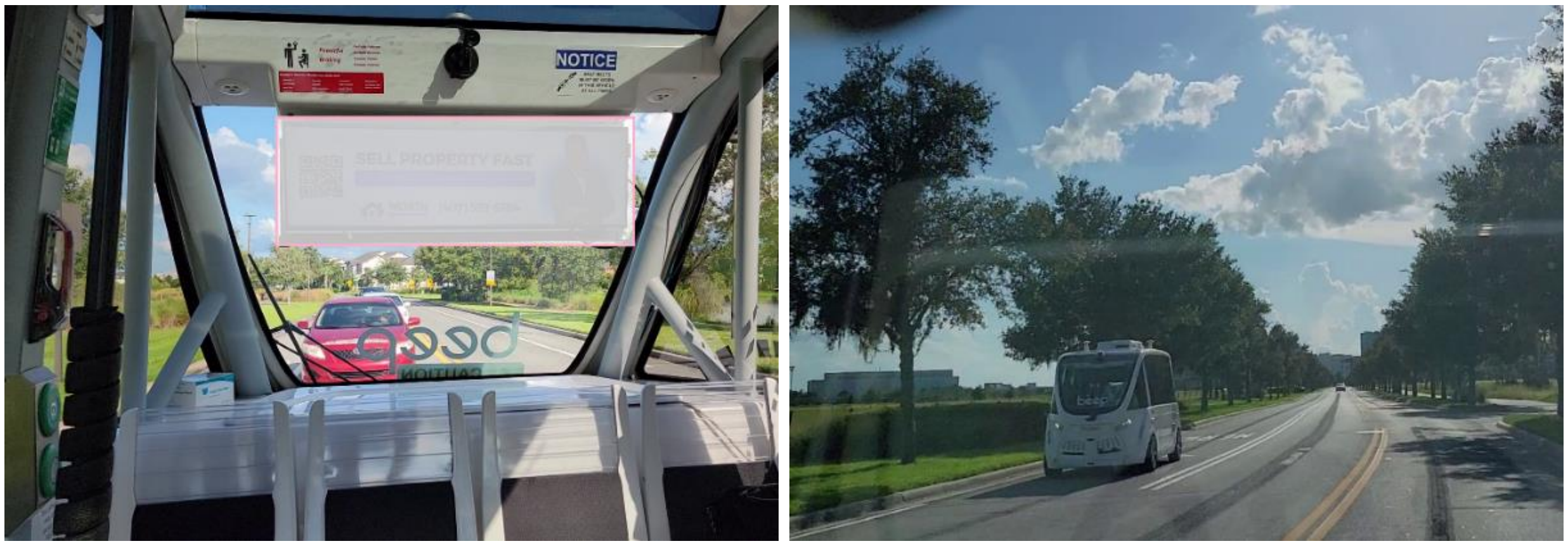

**Figure 3 Photos taken during the trajectory data collection: inside the shuttle, HDV queues are visible behind (Left), and outside the shuttle, here no queue is observed (Right)**

**TABLE 2 Trips taken for vehicle trajectory data collection**

| Vehicle Type | Route | Condition (Following shuttle?) | Number of Trips |
|---|---|---|---|
| HDV | 1 | No | 4 |
| | 1 | Yes | 1 |
| | 3 | No | 2 |
| | 6 | No | 2 |
| Shuttle | 1 | - | 1 |

**TABLE 3 Vehicle modeling parameters**

| Type | Parameter | Unit | Min (HDV) | Mean (HDV) | Max (HDV) | Min (Shuttle) | Mean (Shuttle) | Max (Shuttle) |
|---|---|---|---|---|---|---|---|---|
| Speed | | mph | Governed by Road speed limit | | | 9.5 | 9.5 (0) | 9.5 |
| **Dynamic** | | | | | | | | |
| Main | Speed acceptance | - | 0.9 | 1 (0.25) | 1.2 | 1 | 1 (0) | 1 |
| | Clearance | Meter | 0.5 | 2 (0.5) | 3.5 | 1 | 1 | 1 |
| | Yield time | Second | 5 | 10 (2.5) | 15 | 6 | | |
| Reaction time | Normal | Second | 0.8 | | | 0.1 | | |
| | At stop | Second | 1.3 | | | 0.1 | | |
| | At traffic | Second | 1.7 | | | 0.1 | | |
| **Microscopic** | | | | | | | | |
| Main | Max. acceleration | $m/s^2$ | 2 | 5 (0.5) | 6 | 3 | 3 (0) | 3 |
| | Norm. deceleration | $m/s^2$ | 2.5 | 3 (.5) | 3.5 | 2 | 2 (0) | 2 |
| | Max deceleration | $m/s^2$ | 4 | 5 (.5) | 6 | 6 | 6 (0) | 6 |
| Car-following | Sensitivity | - | 1 | | | 0.3 | 0.7 (0.3) | 0.9 |
| | Gap | - | 0 | | | 2 | | |
| | Aggressiveness level | - | 0 | | | 0 | | |
| Lane-changing | Staying at overtaking lane | - | No | | | No | | |
| | Imprudent lane changing | - | No | | | No | | |
| | | | HDV (Free movement) | | | Shuttle | | |
| **Accuracy** | MAPE of travel time in simulation | | 27.5 | | | 5.8 | | |

**Validation**
The validation of this process is done by comparing the travel time of HDV following the shuttle in the route 1 segments observed in the simulation experiment against the actual time observed in the field experiment. The validation result is shown in **Table 4**. The validation result implies that the absolute error percentage in calibration and validation results for HDV movement are close.

**TABLE 4 Validation of the vehicle modeling (HDVs travel time in route 1 following the shuttle)**

| **Segments on Route 1** | **Observed travel time (s)** | **Mean simulated travel time (s)** | **Absolute Percentage Error** |
|---|---|---|---|
| 1 | 59 | 38.7 | 34.4 |
| 2 | 80 | 109.4 | 38.7 |
| 3 | 201 | 187.7 | 6.6 |
| 4 | No following vehicle is found in simulation | | |
| 5 | 128 | 107.1 | 16.4 |
| 6 | 58 | 41.4 | 28.6 |
| MAPE | | | 24.5 |

**Travel Demand and Route Choice Calibration**
The O-D demands are introduced in the simulation through a centroid-to-centroid demand matrix. We have considered the Central Florida Regional Planning Model (CFRPM) version 7 data for initial demand generation and attraction and defining the centroids (*41*). The study area is in the zip code 32827 where 14 traffic analysis zones (TAZs) are located. These TAZs are considered as demand centroids, which will generate the internal-to-internal (the trips which generate and end inside the area) demand of the study area. We have also considered other two different demand sources, i.e., external-to-external (the trips which generate and end outside of the area but using the study area's road network), and external-to-internal/internal-to-external (the trips whose one end is outside, and the other end is inside) and for this we considered another 4 external centroids, for generating and attracting demand outside of this area. In total, 18 centroids are considered here for O-D matrix (**Figure 3 (Left)**).
The next step is the calibration of the generated demand. The O-D demands are calibrated using aggregated vehicle volume data collected from road detectors. We didn't find any detector inside the Lake Nona community but there are detectors available from Regional Integrated Transportation Information System (RITIS) in Central Florida Greeneway (SR 417) (*42*). In the study area segment, we found 40 detectors which we used for calibration. We calculated the mean detector traffic count in 5-minutes intervals throughout the day, for the year 2022. Next, we adjusted the demand matrix as the CFRPM data is for the base year 2015. The adjustment is done using static O-D adjustment procedure in Aimsun Next (*43*). The trip demand throughout the day in **Figure 3 (Right)** shows there are two peaks in the full day demand. This study considers one-hour off peak (11 AM to 12 PM) and one-hour peak (5 PM to 6 PM) periods for impact assessment.
The O-D adjustment procedure is followed by the mesoscopic simulation which is used to generate path assignment for microscopic simulation. The assigned paths increase the microscopic simulation accuracy. A certain proportion of the vehicles in microscopic simulation have the option to follow the path assigned by mesoscopic simulation and the others can fix their path in microscopic simulation to reach their destination. This proportion is determined through the calibration procedure. The route choice calibration process is done using a dynamic traffic assignment procedure to minimize the difference between actual and simulated data in different volume detectors. The calibrated volume of cars in detectors during off peak and peak periods are given in **Figure 4** left and right, respectively. Calibration accuracy is measured using the GEH value. If the GEH value is less than 5, it indicates very good calibration (*24*, *44*). The calibrated parameters of dynamic traffic assignment are provided in **Table 5**.

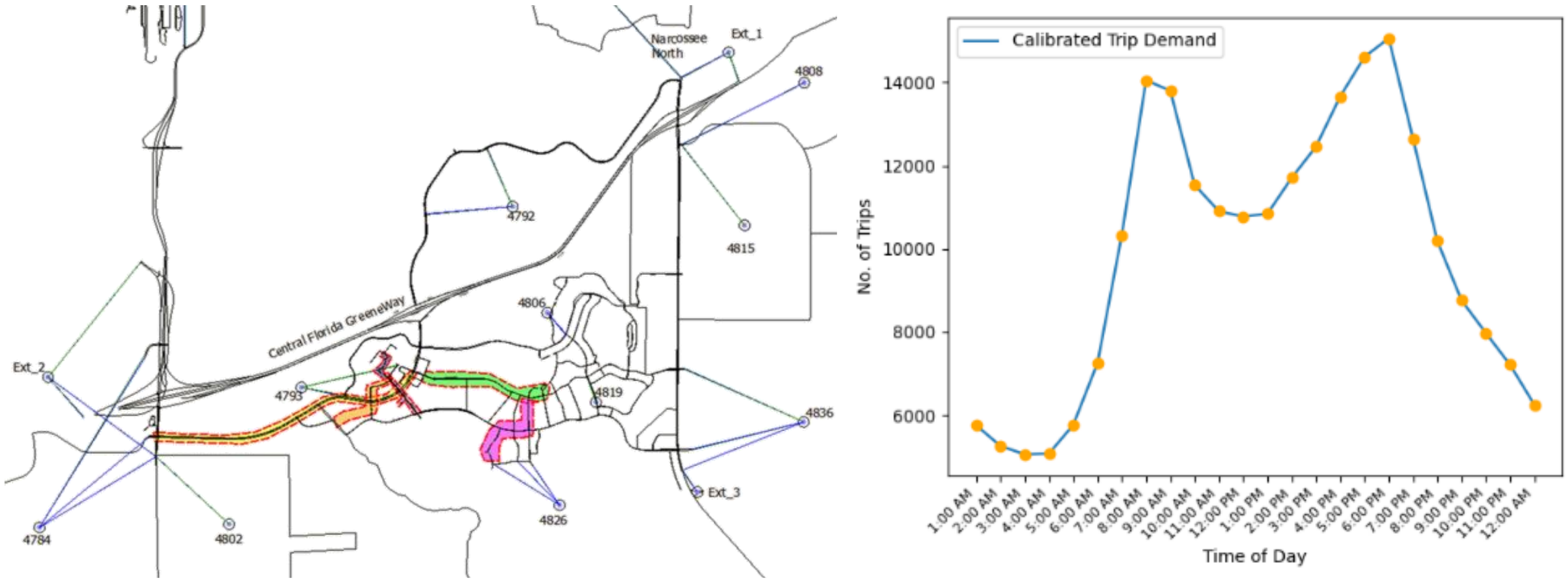

**Figure 3 Trip demand in study area: Traffic generation and attraction centroids (Left) and Trip demand throughout the day (Right)**

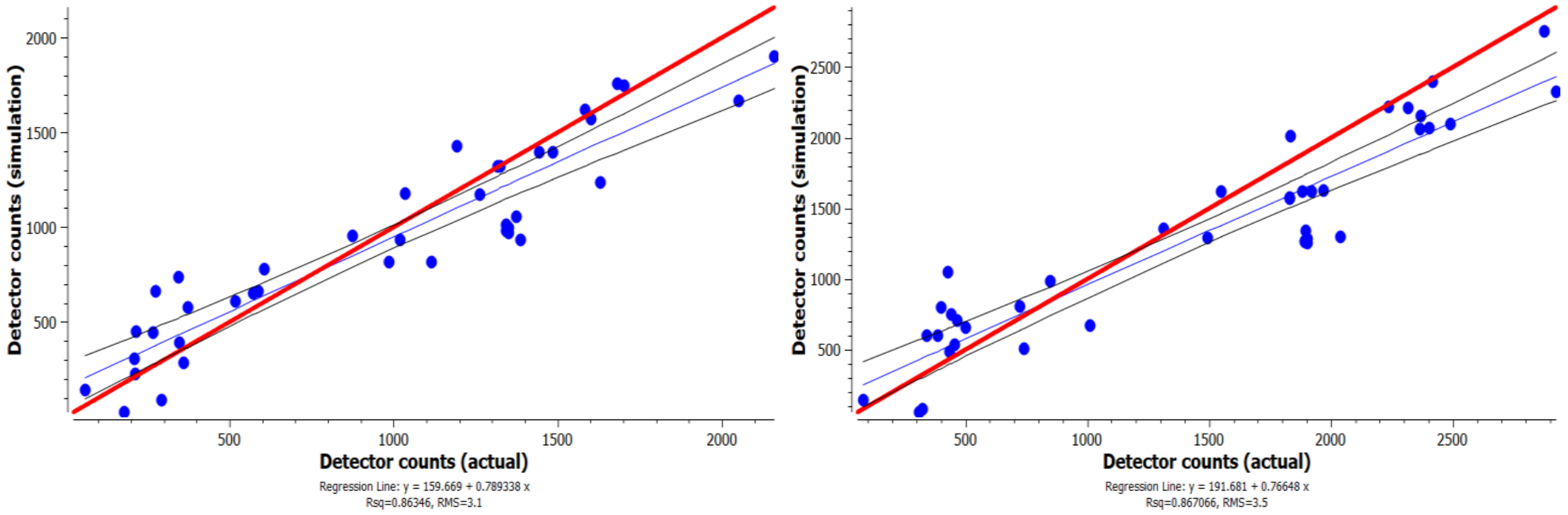

**Figure 4 Detector counts (simulation vs. actual) during off peak period (left) and peak period (right)**

**TABLE 5 Calibration of microscopic simulation through Dynamic Traffic Assignment**

| **Factors** | **Off peak period** | **Peak period** |
|---|---|---|
| Attractiveness weight | 3 | 3 |
| Trips following path assignment from mesoscopic simulation | 70% | 70% |
| Stochastic route choice model | Logit | C-Logit |
| Model parameters | Scale = 12 | Scale = 12, Beta = 0.1, Gamma = 1 |
| Maximum number of initial paths to consider | 3 | 3 |
| Maximum paths per interval | 3 | 3 |
| **Accuracy** | | |
| GEH < 5 | 45.0% | 30.0% |
| GEH < 10 | 75.0% | 65.0% |

### Scenarios for Simulation Experiments

Since our objective is to assess the impact of the shuttle movement on transportation facilities, the best way of analyzing the impact is to measure how the frequency of shuttle movement is contributing to increasing/decreasing the delay in the operating routes and how the traveling speed is being affected. Thus, we have used delay/travel time ratio (%) and speed (mph) of the HDVs as impact assessment metrics. We have considered different scenarios of traffic operations to compare the performance of shuttle movements. We have assessed the impact of shuttles on three routes (No. 1, 3 and 6), the same routes which are used for calibration earlier. The scenarios are as follows:

1. S0: The baseline scenario without any autonomous shuttle movement in all three routes, only HDVs are operated;
2. S1: Long interval (30 minutes) between consecutive shuttles;
3. S2: Medium interval (20 minutes) between consecutive shuttles;
4. S3: Small interval (10 minutes) between consecutive shuttles;
5. S4: Repeat of the S3 scenario, but the shuttle speed is adjusted on trial-and-error basis until the simulation result becomes close to the S0 scenario.

The speed of the HDVs is controlled by road speed limit as well as the vehicle microscopic properties. Two sets of road segments, on both directions, for each of the three routes are considered for impact assessment calculation. The six segment groups are shown in **Figure 5**.

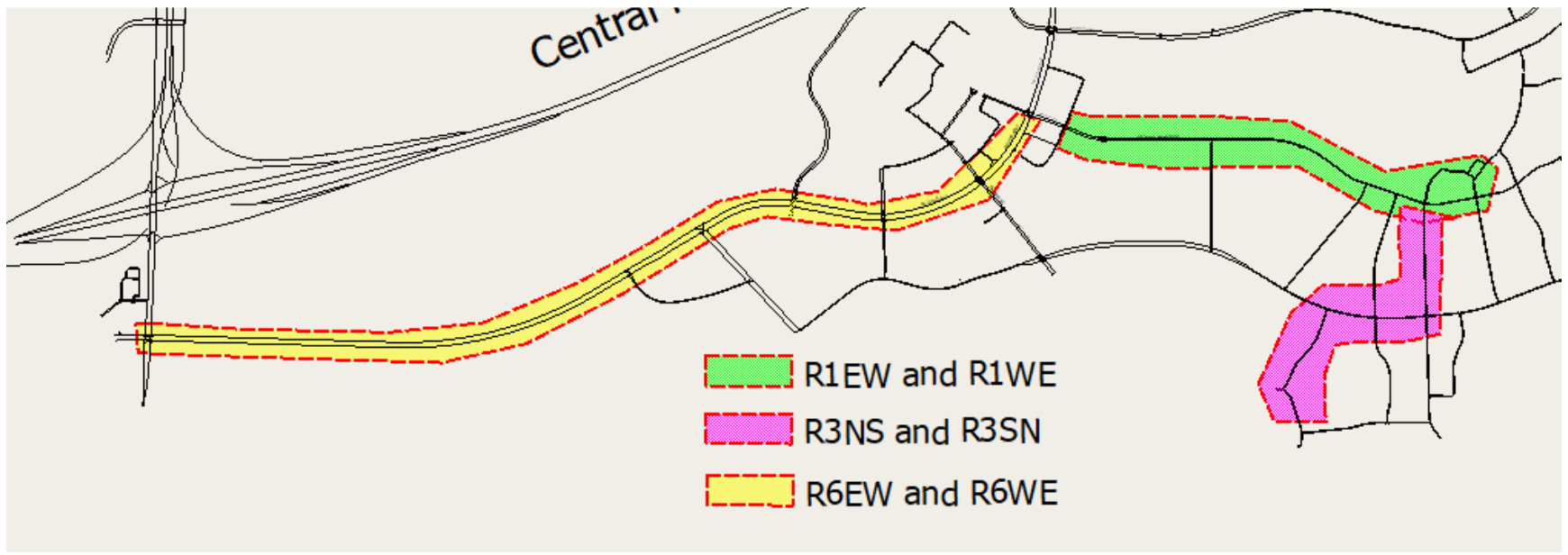

**Figure 5 Position of road segment groups**

**RESULTS**

**Table 6** presents the outputs of the simulated traffic experiments. It is observed that for all the evaluated scenarios the aggregated delay/travel time ratio is higher, and the mean travel speed is lower in peak period, though the difference is not much.

From the implemented scenarios S0 to S3 it is observed that for the HDVs, the aggregated delay/travel time ratio of all the road segment groups increases with the decreasing intervals between shuttles. The delay/travel time ratio is increased by 3.22% in the off-peak and by 3.58% in peak period from scenario S0 to S3. The weighted speed of the traffic flow also reduces by 1.61 mph in the off-peak period modeling and by 2.01 mph in peak period modeling from scenario S0 to S3. The reason for this increase in delay time and decrease in speed is probably the difference in speed between the shuttles and HDVs. The road speed limit in route 1 is 25 mph, in route 3 it varies between 20 and 15 mph in different sections, and for route 6 the speed limit is 35 mph. The HDV generally run close to the road speed limit, but the shuttle speed is calibrated to 9.5 mph in this study. Most of the sections in routes 1 and 3 are one lane only, which means no overtaking or limited overtaking scope for the queuing vehicles behind the shuttle. The delay/travel time ratio is considerably high in the segment groups of route no. 6 due to presence of a number of signalized intersections in that route. The other two routes don't have any signalized intersection; thus, the delay is comparatively less on these routes due to low traffic volume.

It is observed that the increase in delay/travel time ratio and the decrease in speed with shuttle interval reduction is not consistent, i.e., some segment groups observe decrease in delay/travel time ratio or increase in speed due to shuttle interval reduction. The probable reason behind this might be the change in route choice algorithm by the vehicles during the simulation to reduce delay.

**TABLE 6 Traffic statistics of HDV's from simulation in different scenarios**

| Scenarios (→) Parameters (→) Segment groups (↓) | S0 | | S1 | | S2 | | S3 | | S4 | |
|---|---|---|---|---|---|---|---|---|---|---|
| | Delay/ Travel Time (%) | Speed (mph) | Delay/ Travel Time (%) | Speed (mph) | Delay/ Travel Time (%) | Speed (mph) | Delay/ Travel Time (%) | Speed (mph) | Delay/ Travel Time (%) | Speed (mph) |
| **Off peak period** | **Shuttle speed 9.5 mph** | | | | | | | | **Shuttle speed 15 mph** | |
| **R1EW** | 5.09 | 25.02 | 7.35 | 24.53 | 9.92 | 23.03 | 14.6 | 21.10 | 5.63 | 23.81 |
| **R1WE** | 5.12 | 23.11 | 7.05 | 23.38 | 8.23 | 23.24 | 10.29 | 21.99 | 6.48 | 24.51 |
| **R3NS** | 1.64 | 15.39 | 3.21 | 14.95 | 4.98 | 14.74 | 2.52 | 14.47 | 2.60 | 16.40 |
| **R3SN** | 10.77 | 15.33 | 11.89 | 13.26 | 13.42 | 13.21 | 13.98 | 12.71 | 11.33 | 13.18 |
| **R6EW** | 28.34 | 23.25 | 30.78 | 22.63 | 24.78 | 21.21 | 28.5 | 24.84 | 26.85 | 23.39 |
| **R6WE** | 23.38 | 24.59 | 24.19 | 25.36 | 19.97 | 23.50 | 18.37 | 22.92 | 21.58 | 25.42 |
| **Aggregated** | **11.47** | **21.97** | **13.02** | **21.65** | **13.25** | **20.75** | **14.69** | **20.36** | **11.61** | **22.03** |
| **Peak period** | **Shuttle speed 9.5 mph** | | | | | | | | **Shuttle speed 20 mph** | |
| **R1EW** | 5.82 | 24.94 | 9.51 | 24.15 | 10.92 | 23.59 | 13.16 | 23.54 | 6.32 | 24.84 |
| **R1WE** | 6.72 | 24.55 | 7.83 | 24.46 | 10.49 | 23.00 | 12.14 | 20.91 | 7.14 | 24.02 |
| **R3NS** | 2.46 | 15.79 | 4.41 | 15.32 | 5.3 | 15.07 | 5.26 | 14.41 | 3.38 | 16.38 |
| **R3SN** | 12.12 | 12.81 | 12.54 | 12.59 | 13.86 | 11.92 | 14.84 | 11.76 | 12.2 | 12.37 |
| **R6EW** | 29.46 | 23.06 | 28.94 | 22.29 | 28.4 | 22.04 | 28.83 | 19.67 | 30.12 | 23.98 |
| **R6WE** | 28.33 | 24.8 | 28.19 | 24.41 | 26.81 | 25.20 | 28.49 | 24.58 | 26.86 | 25.11 |
| **Aggregated** | **13.17** | **22.05** | **14.48** | **21.59** | **15.41** | **21.08** | **16.75** | **20.04** | **13.38** | **22.10** |

Scenario S4 is created to test the hypothesis that increasing shuttle speed will improve the road condition by decreasing delays. It is observed that when the shuttle speed is increased to 15 mph during the off-peak period, the average delay/travel time ratio for HDV is reduced from 14.69% to 11.61%, close to the baseline scenario S0 with no shuttle, despite the shuttle frequency remaining at 10 min/trip. The weighted mean speed of the road segments also improved from 20.36 mph in S3 to 22.03 mph in S4. Similar improvements are also observed during the peak period when the shuttle speed increases to 20 mph. Nevertheless, this is only a hypothetical solution to this issue. Increasing the speed of autonomous shuttles depends on many factors including prevailing regulations to ensure safety of other road users such as drivers and pedestrians etc.

## CONCLUSIONS

Autonomous vehicles are expected to transform urban transportation with enormous opportunities offered by this technology. In this regard, autonomous shuttles are supposed to bring revolutionary changes in public transportation. This present study focuses on assessing the impact of autonomous shuttle movement on urban road networks through calibration of a microscopic simulation model using real-world trajectory data. For this assessment, an operational shuttle system at Lake Nona, Orlando is emulated in the simulation. One of the contributions of this study is that the microscopic behaviors of the operating shuttles, HDVs, and their interactions are calibrated through vehicle trajectory data collected from filed experiments. This study implemented several scenarios with different frequencies of shuttle movement along with calibrated total trip demand in the analyzed area transportation model. Comparing results from different scenarios suggests that a low operating speed of autonomous shuttles can create bottlenecks and therefore increase delays and reduce traffic flow.

There are several outcomes of this study. First, this study calibrates movement of autonomous vehicles using real-world trajectory data; the calibrated parameters can be used in future for simulating the microscopic behavior of automated shuttles. These parameters can accurately represent autonomous vehicles in simulation, which would be very important to evaluate their traffic and safety impacts. Second, this study evaluates the limits of an existing autonomous shuttle system and provides insights on how to improve system performance.

There are several limitations of the study. First, the internal traffic demand of Lake Nona area is not available. As a result, we had to distribute the incoming and outgoing traffic from/to Lake Nona based on some assumptions. Second, the signal plans of the intersections in the Lake Nona Boulevard are not available, which made us establish the signal timings based on some assumptions. Third, we haven't considered the reduction in traffic demand inside the Lake Nona area due to shuttle movement. We assumed that the changes in traffic demand are negligible since the shuttle system is not fully operational.

Future studies in this direction can also incorporate different penetration levels for shuttles by adding more routes in the study area to observe the impact and to better prepare us for a future where current vehicle movement is combined with autonomous vehicles and shuttles.

**ACKNOWLEDGEMENT**

The authors would like to thank Mr. Glenn Cook, founder and CEO of EVTransports and Mr. Mark Reid, Co-founder of Beep for their valuable information regarding Beep shuttle movement and facilitating field data collection; and Mr. Mobasshir Rashid of UCF for his valuable efforts in vehicle trajectory data collection. The authors would also like to acknowledge the partial support from Mobility Insight Ltd.

**AUTHOR CONTRIBUTIONS**

The authors confirm contribution to the paper as follows: study conception and design: S. Roy, B. Nahmias-Biran, S. Hasan; data collection: S. Roy; analysis and interpretation of results: S. Roy, B. Nahmias-Biran, S. Hasan; draft manuscript preparation: S. Roy, S. Hasan. All authors reviewed the results and approved the final version of the manuscript.